\documentclass{article}

\usepackage[nonatbib,final]{neurips_2020}
\usepackage[utf8]{inputenc} 
\usepackage[T1]{fontenc}    
\usepackage{subfig,hyperref,url,booktabs,amsfonts,nicefrac,microtype,amsmath,amsthm,amssymb,algorithmic,graphicx,xcolor,wrapfig,lipsum,mathtools,algorithm,caption,multirow,wrapfig}
\hypersetup{colorlinks=true,linkcolor=blue,citecolor=red}
\graphicspath{{figures/}}

\theoremstyle{definition}

\DeclareMathOperator{\tr}{Tr}
\DeclareMathOperator{\Tr}{Tr}

\renewcommand{\a}{{\bf a}}

\newcommand{\n}{{\bf n}}

\newcommand{\x}{{\bf x}}
\newcommand{\y}{{\bf y}}
\newcommand{\z}{{\bf z}}

\newcommand{\C}{{\bf C}}

\newcommand{\F}{{\bf F}}

\newcommand{\I}{{\bf I}}

\newcommand{\M}{{\bf M}}

\newcommand{\Q}{{\bf Q}}
\newcommand{\R}{\mathbb{R}}

\newcommand{\U}{{\bf U}}

\newcommand{\W}{{\bf W}}
\newcommand{\X}{{\bf X}}
\newcommand{\Y}{{\bf Y}}

\newcommand{\Sig}{\boldsymbol{\Sigma}}

\title{A biologically plausible neural network\\for local supervision in cortical microcircuits}

\author{Siavash Golkar$^{\,1}$    \hspace{25pt}   David Lipshutz$^{\,1}$  \hspace{25pt}   Yanis Bahroun$^{\,1}$ \vspace{7pt}
   \\
   \textbf{Anirvan M.\ Sengupta$^{\,1,2}$ \hspace{25pt} Dmitri B.\ Chklovskii$^{\,1,3}$  }
  \vspace{14pt}
   \\
   $^{1\,}$Center for Computational Neuroscience, Flatiron Institute
   \\
   $^{2\,}$Department of Physics and Astronomy, Rutgers University
   \\
   $^{3\,}$Neuroscience Institute, NYU Medical Center
   \vspace{5pt}\\
   \texttt{\{sgolkar,dlipshutz,ybahroun,mitya\}@flatironinstitute.org}\\
   \texttt{anirvans.physics@gmail.com }
}
\begin{document}
\maketitle

\begin{abstract}
The backpropagation algorithm is an invaluable tool for training artificial neural networks; however, because of a weight sharing requirement, it does not provide a plausible model of brain function. Here, in the context of a two-layer network, we derive an algorithm for training a neural network which avoids this problem by not requiring explicit error computation and backpropagation. Furthermore, our algorithm maps onto a neural network that bears a remarkable resemblance to the connectivity structure and learning rules of the cortex.  We find that our algorithm empirically performs comparably to backprop on a number of datasets.
\end{abstract}

\section{Introduction}\label{sec:intro}
The Backpropagation algorithm~\cite{Rumelhart1986} (backprop) is an invaluable tool for machine learning, which has allowed for the efficient training of deep neural networks. However, despite its effectiveness, backprop is not biologically plausible~\cite{weight_transport_problem1,weight_transport_problem2,weight_transport_problem3}. In particular, in backprop, the weights used for computing the forward pass of the network are also used in the backward pass, leading to an implausible weight sharing in the circuit referred to as the weight transport problem~\cite{weight_transport_problem1}.

To address this issue, a number of modifications to backprop have been proposed~\cite{Burbank2012,Burbank2015,Lillicrap2016,Akrout2019}. New learning algorithms based on predictive coding~\cite{Lee2014} and target propagation~\cite{Whittington2017}, and empirically based learning rules~\cite{Sacramento2018, Payeur2020} have also been discussed. However, these works generally suffer from either performance issues or do not account for experimentally known properties of the cortex (see~\cite{Lillicrap2020} for a recent review).

In this workshop abstract, in the context of a two-layer neural network, we derive an algorithm that does not suffer the plausibility problems of backprop. Specifically, adding gain control by imposing an inequality constraint results in an algorithm that avoids the weight sharing problem by not explicitly computing and backpropagating the error. Furthermore, we find that the resulting neural network bears remarkable resemblance to cortical connectivity structures and approximates the calcium plateau dependent learning rules in pyramidal neuron synapses.
{\let\thefootnote\relax\footnote{{This abstract, presented at the NeurIPS 2020 workshop ``Beyond Backpropagation'',  is based on the recent work~\cite{golkar2020} and highlights the relationship between the proposed algorithm and backpropagation.}}

\section{A two-layer linear network}

In this section, we introduce an objective function with a quadratic loss and demonstrate how it can be solved by a linear neural network using backprop. We then show how the addition of a gain control constraint leads to a linear neural network with biologically plausible (i.e., local) learning rules. While we present the problem in the context of a linear network, as we discuss in Sec.~\ref{sec:discussion}, our approach naturally extends to the nonlinear setting.

Given inputs $\x_1,\dots,\x_T$ and labels $\y_1,\dots,\y_T$, with $\x_t\in\R^m$ and $\y_t\in\R^n$, consider the objective:
\begin{equation}\label{eq:lin_obj}
    \min_{\W_1,\W_2} \frac1T\sum_t \|\y_t-\W_2 \W_1 \x_t\|^2,
\end{equation}
where $\W_1\in\R^{k\times m}$ (resp. $\W_2\in\R^{n\times k}$) are the weights of the first (resp. second) layer of the network. We define $\z_t := \W_1 \x_t$ as the $k$-dimensional activities of the hidden layer and $\hat \y_t := \W_2 \W_1 \x_t$ as the network's prediction of the label $\y_t$ given input $\x_t$. This linear problem can be thought of as a rank-constrained linear regression problem.

\subsection{Training with backpropagation}

When trained by backprop, the weight updates of this network are given by taking derivatives of the loss with respect to the weights~\cite{SGD}. Here, we focus on the learning rule for the weights of the first layer:
\begin{equation}\label{eq:backprop_W1}
    \delta \W_1 \propto (\W_2^\top \epsilon_t)\x_t^\top \quad,\quad \epsilon_t = (\y_t-\hat \y_t),
\end{equation}
where we have defined $\epsilon_t$ as the prediction error for the sample at time $t$. A cartoon of the process for computing the update for $\W_1$ is given in Fig.~\ref{fig:rrr_ann}, where the forward and backward passes are respectively denoted in blue and red.  Here, the problem of weight transport  is manifest in the use of the weights $\W_2$ both in the forward pass when computing the error $\epsilon_t$, and also their transpose in the backward pass when propagating the error back to the first layer.

\begin{figure}[ht]
\vspace{-10pt}
\centering
\subfloat[\small two-layer artificial neural network] {\label{fig:rrr_ann}\includegraphics[width=0.47\textwidth]{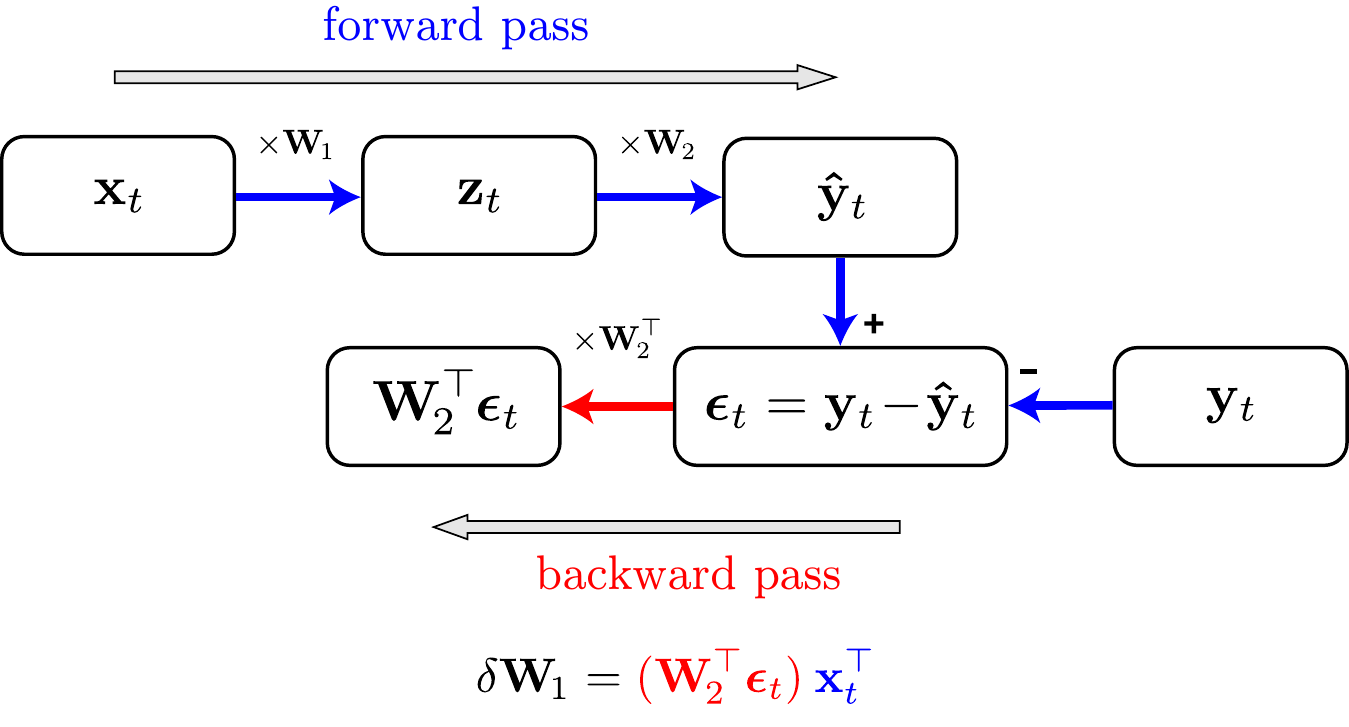}}
\hfill
\subfloat[\small BMVR algorithm]{\label{fig:rrr_bio}\includegraphics[width=0.47\textwidth]{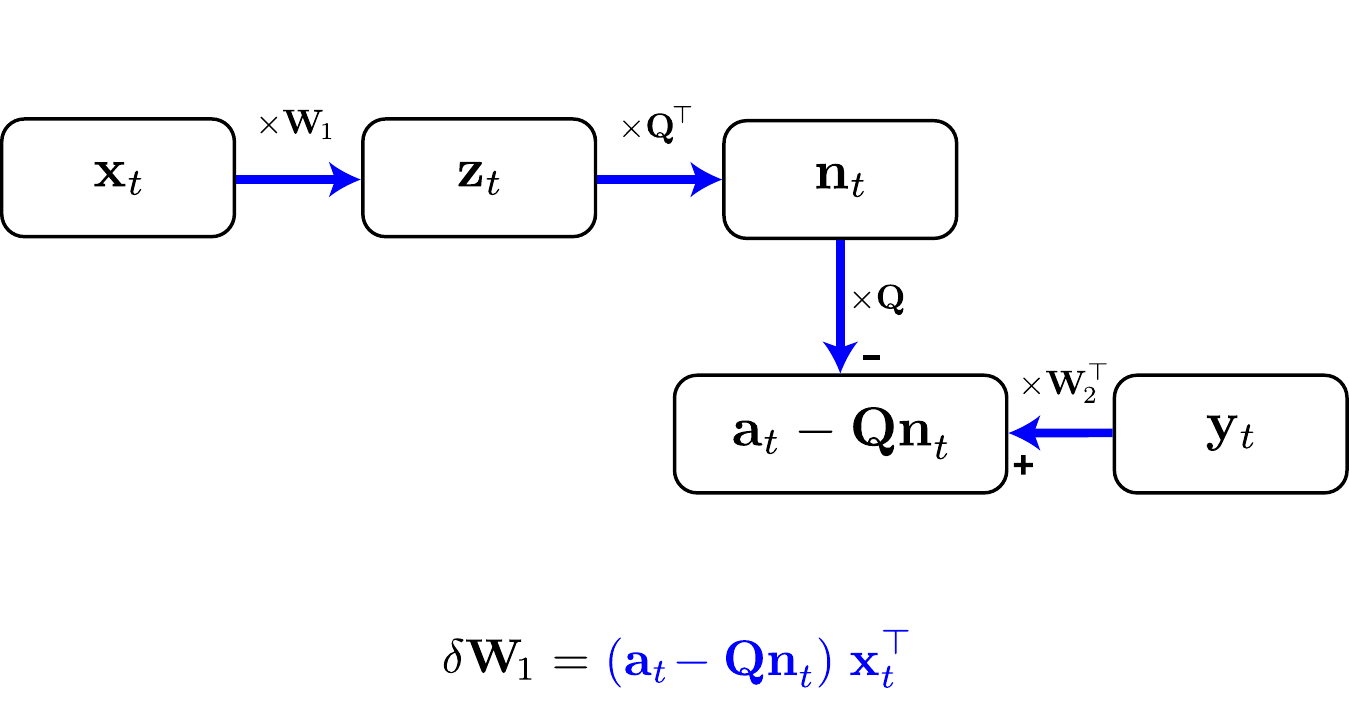}}
\caption{\small (left) Schematic of a two-layer linear neural network trained by backprop, demonstrating the computation of the learning rule for the weights of the first layer $\W_1$. The blue and red arrows respectively denote the forward and backward passes. (right)  Schematic of BMVR, demonstrating the computation of the learning rule for $\W_1$.
In BMVR, $\color{blue}\a_t-\Q\n_t$ (encoded in the calcium plateau potential) replaces the backpropagated error $\color{red}\W_2^\top \boldsymbol \epsilon_t$ in the $\W_1^\top$ learning rule. } \end{figure}

\subsection{A biological network with gain control}
We now show that adding a gain control mechanism to the objective given in Eq.~\eqref{eq:lin_obj} unexpectedly results in a biologically plausible neural network without backprop. We implement this gain control mechanism by requiring the activity of the hidden layer to satisfy $\tfrac1T\sum_t\z_t\z_t^\top\preceq \I_k$, i.e., \mbox{$\I_k-\tfrac1T\sum_t\z_t\z_t^\top$} is a positive semi-definite matrix. This inequality can be incorporated into the objective function by adding a Lagrange multiplier to Eq.~\eqref{eq:lin_obj}:
\begin{equation}\label{eq:lin_obj_const}
\min_{\W_1,\W_2}\max_{\Q}\frac1T\sum_t \|\y_t-\W_2 \W_1 \x_t\|^2 +\Tr \Q\Q^\top(\W_1\x_t\x_t^\top \W_1^\top - T\times\I_k),
\end{equation}
where $\Q\Q^\top$ is a positive-definite matrix playing the role of the Lagrange multiplier~\cite{pehlevan2015normative}. Expanding the square term and using the inequality constraint we arrive at an upper bound for the objective:
\begin{multline}\label{eq:lin_obj_UB}
    \eqref{eq:lin_obj_const}\;\leq\;\min_{\W_1,\W_2}\max_{\Q} \frac1T\sum_t \Big[ \y_t^\top\y_t-2\y_t^\top\W_2 \W_1 \x_t+\Tr \,\W_2\W_2^\top 
    \\ 
    + \Tr\Q\Q^\top(\W_1\x_t\x_t^\top \W_1^\top - T\times\I_k)\Big].
\end{multline}
In Sec.~\ref{app:const_saturation} of the supplementary materials, we show that this inequality is saturated at the global minimum of the objective. Therefore the upper-bound is tight and by optimizing this second objective we find the same optima as we would by explicitly computing and backpropagating the error, \mbox{i.e., \eqref{eq:lin_obj_const} = \eqref{eq:lin_obj_UB}}. We can derive the learning rules for this system by taking stochastic gradient descent-ascent steps. Explicitly, at time $t$ we have:
\begin{align}
    \W_1&\gets\W_1+\eta(\a_t-\Q\n_t)\x_t^\top\label{eq:W1_rule}\\
    \W_2^\top&\gets\W_2^\top+\eta (\z_t\y_t^\top -\,\W_2^\top)\label{eq:W2_rule}\\
    \Q&\gets\Q+\frac{\eta}{\tau}(\z_t\n_t^\top-\Q)\label{eq:Q_rule}.
\end{align}
where  $\z_t:=\W_1\x_t$ is the output of the algorithm, $\a_t:=\W_2^\top\y_t$ and $\n_t:=\Q^\top\z_t$. Here, we have given the update for $\W_2^\top$ instead of $\W_2$ for purposes of biological interpretation (see Sec.~\ref{sec:circuit}). We call this algorithm biological multi-variate regression (BMVR).

A cartoon of the process for computing the update for $\W_1$ is given in Fig.~\ref{fig:rrr_bio}. In this algorithm, the weights $\W_2^\top$ are \emph{only} used to backpropagate the target $\y_t$ and  notably neither the prediction of the network $\hat\y_t$ nor the prediction error $\epsilon_t$ is explicitly computed. In this way, we avoid the weight transport problem. Even though this algorithm does not explicitly compute the error in the forward pass, in the supplementary materials Sec.~\ref{app:bioRRR_interpretation} we show that the quantity $\a_t-\Q\n_t$ can still be interpreted as an implicit backpropagated error signal.
For a more detailed comparison of this algorithm with prior work, in particular with target propagation~\cite{Lee2014}, see supplementary materials Sec.~\ref{app:prior_work}.

\section{Biological implementation and experimental evidence}\label{sec:circuit}

\setlength{\columnsep}{20pt}

\begin{wrapfigure}{r}{0.53\textwidth}
\vspace{-12pt}
\centering
\includegraphics[trim=0 0 0 0, clip,width=0.52\textwidth]{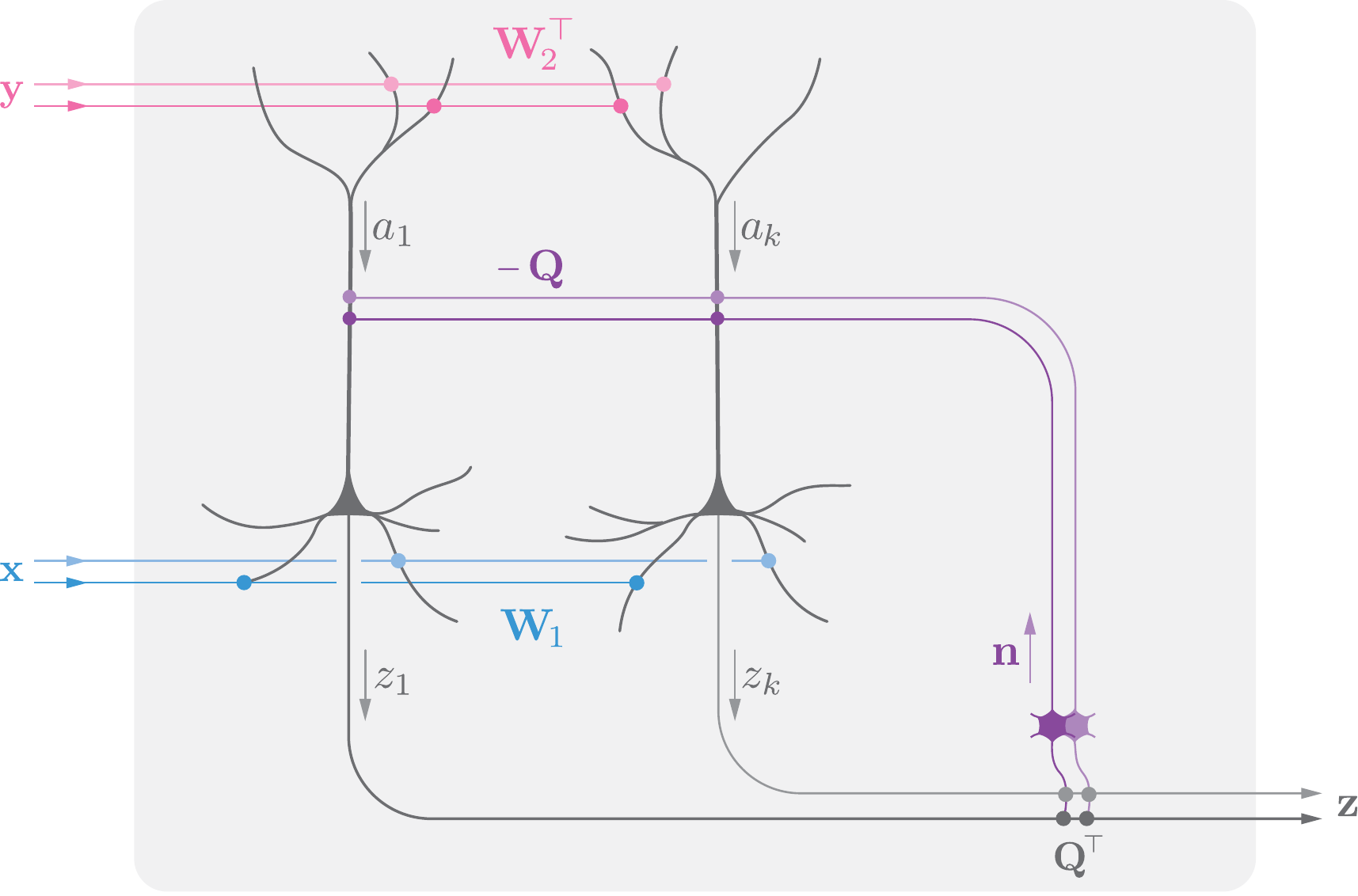}
\caption{\small Cortical microcircuit for BMVR. Pyramidal neurons (black) receive inputs $\x$ onto the dendrites proximal to the cell bodies (black triangles) weighted by $\W_1$, and inputs $\y$ onto the distal dendrites weighted by $\W_2^\top$. 
Output activity of pyramidal neurons, $\z=(z_1,\dots,z_k)$, is fed back via inhibitory interneurons (purple). }
\label{fig:microcircuit}
\vspace{-5pt}
\end{wrapfigure}

The BMVR algorithm summarized by the update rules in Eqs.~\eqref{eq:W1_rule}$-$\eqref{eq:Q_rule} can be implemented in a neural circuit with schematis shown in Fig.~\ref{fig:microcircuit}. In this circuit, the individual components of the output of BMVR, $z_1,\dots, z_k$, are represented as the outputs of $k$ neurons . The matrices $\W_1$ and $\W_2^\top$ are encoded as the synaptic connections between the pyramidal neurons and the inputs of the network (blue and pink nodes in Fig.~\ref{fig:microcircuit}). 
Because of the disjoint nature of the two inputs, we model these as separate dendritic compartments, denoted in Fig.~\ref{fig:microcircuit} as the top (apical tuft) and bottom (proximal) branches of the output neurons. 
The quantities $\z_t=\W_1 \x_t$ and $\a_t=\W_2^\top \y_t$ are then the integrated dendritic currents in each compartment. In the cortex, the role of these output neurons is played by pyramidal neurons which have two distinct dendritic compartments, the proximal compartment comprised of the basal and proximal apical dendrites providing inputs to the soma, and the distal compartment comprised of the apical dendritic tuft~\cite{Larkum756,pyramidal_review}. These two compartments receive excitatory inputs from two separate sources~\cite{takahashimagee,Larkum2013}. 

Similarly, the auxiliary variable $\n$ is represented by the activity of $k$ interneurons
with $\Q$ encoded in the weights of synapses connecting $\n$ to $\z$ (purple nodes on the upper dendritic branch of $\z$) and $\Q^\top$ encoded in the weights of synapses from $\z$ to $\n$ (gray nodes). 
In a biological setting, the implied equality of weights of synapses from $\z$ to $\n$ and the transpose of those from $\n$ to $\z$ can be guaranteed approximately by application of the same Hebbian learning rule (see supplementary materials Sec.~\ref{app:decoupled_weights}). In the cortex, somatostatin-expressing interneurons --- which preferentially inhibit the apical dendrites~\cite{klausberger2003brain,riedemann2019diversity} --- play the role of the $\n$ variables.

The update rule for $\W_1$ (Eq.~\ref{eq:W1_rule}) is an outer product of two terms. The first term $\a_t-\Q\n_t$ is the difference between excitatory synaptic current on the apical tuft ($\a_t = \W_2^\top \y_t$) and inhibitory current induced by interneurons ($\Q\n_t$).
This matches experimental observations that the calcium plateau potential, similarly encoding the difference between the apical tuft current and the interneuron inhibition, drives the plasticity of the proximal synapses~\cite{golding2002,bittner2015,bittner2017behavioral,GreinbergerMagee2020}. 
The plasticity rule for $\W_2$ (Eq.~\ref{eq:W2_rule}) is Hebbian, also matching experimental observations in the neocortex~\cite{sjostrom2006}.

\section{Numerical experiments} \label{sec:experiments}
We implemented our algorithm on a number of standard benchmarks and compared with backprop. The results of the experiments in the linear case in terms of the objective~\eqref{eq:lin_obj} are given in Fig.~\ref{fig:experiment}. We see that in all cases, BMVR has a higher loss during training, but reaches the same optimum with the same number of iterations. This is as expected since the BMVR  objective~\eqref{eq:lin_obj_const} is an upper bound to the objective~\eqref{eq:lin_obj}, which is saturated at the optimum.

\begin{figure}[ht]
\vspace{-5pt}
\hspace{-0.05\textwidth}\includegraphics[width=1.07\textwidth]{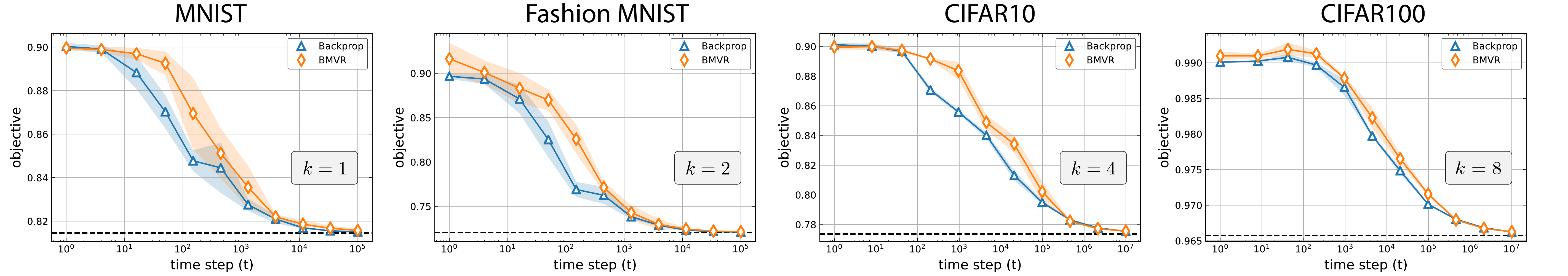}
\caption{\small Comparisons of the backpropagation and BMVR algorithms in terms of the objective Eq.~\eqref{eq:lin_obj} vs. the time-step.  Mean $\pm$ standard deviation over 5 runs of the experiment. }
\label{fig:experiment} 
\vspace{-5pt}
\end{figure}

\section{Discussion}\label{sec:discussion}

We have shown that adding a gain control mechanism via an inequality constraint to a two-layer linear feedforward neural network leads to a micro-circuit which is both biologically plausible and also matches many known properties of pyramidal neurons in the cortex. We showed empirically that the resulting algorithm performs comparably to the backpropagation algorithm on a number of tasks. Here, we consider some generalizations of this work. 

\paragraph{Non-linear extension.} The extension of our algorithm to networks with non-linearity is straightforward. Starting from the objective:
\begin{equation}\label{eq:nonlin_obj}
    \min_{\W_1,\W_2} \frac1T\sum_t \|\y_t-\W_2 f(\W_1 \x_t)\|^2,
\end{equation}
we again impose an inequality constraint on the activity of the hidden layer $\z_t := f(\W_1 \x_t)$ such that  $\tfrac1T\sum_t\z_t\z_t^\top\preceq \I_k$. Following the same steps, we derive identical update rules for $\W_2$ and $\Q$ and a slightly modified update for $\W_1$:
\begin{align}
    \W_1&\gets\W_1+\eta f'(\z_t)(\a_t-\Q\n_t)\x_t^\top.\label{eq:W1_NN_rule}
\end{align}
In this case, the proof of the tightness of the inequality no longer holds in general. When the non-linearity is ReLU, $f'(\z_t)$ in the above expression is replaced by $(\z_t>0)$ 
such that the first layer weights are only updated if the hidden layer neuron they are attached to is active (i.e. $W_1^{ij}$ is updated only if $z^i_t>0$). In this way, the non-linear extension maintains and enhances the biological plausibility of the algorithm.

\begin{table}
\centering
\vspace{-5pt}
\scalebox{0.98}{
  \begin{tabular}{ccccc}
    \toprule
    \multirow{2}{3em}{Method} &\multicolumn{2}{c}{$k=64$}  &    \multicolumn{2}{c}{$k=16$}                   \\
    & Train     & Test     & Train & Test \\
    \midrule
    Backprop & 99.8\%  & 94.0\%   & 100\% & 95.5\% \\
    BMVR     & 99.3\% & 93.6\% & 100\% & 95.2\%\\
    \bottomrule
  \end{tabular}}
  \vspace{6pt} 
\caption{\small Comparing the performance of the backprop and BMVR algorithms on a two-layer network with ReLU non-linearity on the MNIST dataset.}
\label{tab:nn}
\vspace{-5pt}
\end{table}
The performance of the algorithm with the non-linear extension for MNIST classification for two different dimensions of the hidden layer $k=16$ and $k=64$ can be seen in Tab.~\ref{tab:nn}. Here, again we achieve performance comparable to that of backprop.  The details of these numerical experiments are given in Sec.~\ref{app:exp_details} of the supplementary materials.

\paragraph{Deep extension.} In this workshop abstract, we focused on a biologically plausible two-layer network which maps onto a micro-circuit in one region of the cortex. By hierarchically combining such networks in a sequence of layers, we can emulate the behavior of deep neural networks. This direction of research is beyond the scope of this abstract and is currently under investigation.

\section*{Acknowledgments}
We are grateful to Jeffrey Magee and Jason Moore for insightful discussions related to this work. We further thank Nicholas Chua, Shiva Farashahi, Johannes Friedrich, Alexander Genkin, Tiberiu Tesileanu, and Charlie Windolf for providing feedback on the manuscript.

\bibliography{biblio.bib}
\bibliographystyle{unsrt}

\clearpage

\appendix

\begin{center}\LARGE{\textbf{Supplementary Materials}}\end{center}

\section{Saturation of the BMVR inequality constraint}\label{app:const_saturation}
Here we show that the inequality constraint $\tfrac1T\sum_t\z_t\z_t^\top\preceq \I_k$ imposed in BMVR is saturated at its optimum in the offline setting. This was previously shown in~\cite{pehlevan2015normative}. Here we provide an alternative proof. The optimization objective is given in Eq.~\eqref{eq:lin_obj_UB} which for brevity we rewrite as
\begin{equation}
    \min_{\W_1\in\R^{k\times m}}\min_{\W_2\in\R^{n\times k}}\max_{\Q\in\R^{k \times k}} \tr
    \W_2^\top\W_2-2\W_1\C_{xy}\W_2^\top+\Q\Q^\top(\W_x\C_{xx}\W_x^\top-\I_k),
\end{equation}
where we have dropped the term $\y_t^\top\y_t$ which is independent of the optimizaiton varaibles and defined the correlation matrices:
\begin{equation}
\C_{xy} = \frac1T\sum_t \x_t\y_t^\top\;,\; \C_{xx} = \frac1T\sum_t \x_t\x_t^\top.
\end{equation}
Here we will assume that $\C_{xx}$ is full rank and $\C_{xy}$ has at least $k$ non-zero eigenvalues.\footnote{If $\C_{xx}$ is not full rank, the same proof would go through by first projecting $\x_t$ on its full rank subspace.} We first find the optimum for $\W_2$ by setting the $\W_2$ derivative to zero:
\begin{equation*}
    0=\W_1 \C_{xy}-\W_2\;\Rightarrow\; \W_2 = \W_1 \C_{xy}.
\end{equation*}
Plugging this back into the optimization objective yields
\begin{equation}
    \min_{\W_x}\max_{\Q} 
    \tr -\W_1\C_{xy}\C_{yx}\W_1^\top+\Q\Q^\top(\W_1\C_{xx}\W_1^\top-\I_k).\label{eq:VxQ_obj}
\end{equation}
The equilibrium condition for this system is given by setting gradients with respect to $\W_1$ and $\Q$ to zero which gives:
\begin{align}
    0&=\W_1 \C_{xy} \C_{yx}-\Q\Q^\top\W_1\C_{xx},\label{eq:Vx_update_apdx}\\
    0&=\Q^\top(\W_1^\top\C_{xx}\W_1^\top-\I_k),\label{eq:Q_update_apdx}
\end{align}
Note that Eq.~\eqref{eq:Q_update_apdx} on its own does not imply that $\W_1\C_{xx}\W_1^\top=\I_k$. However, if we can prove that $\Q$ which is a $k\times k$ matrix, is full rank and has no zero eigenvalues, then Eq.~\eqref{eq:Q_update_apdx} implies $\W_1\C_{xx}\W_1^\top=\I_k$. This is a realization of the fact that when imposing an inequality constraint, for example $f(x)>0$, via a Lagrange multiplier $\lambda$ by optimizing $\min_x \max_{\lambda\geq0} \lambda f(x)$, if the Lagrange multiplier at the optimum is slack $\lambda>0$, then the inequality constraint is saturated $f(x) = 0$. 

In what follows we show that at equilibrium, $\Q\Q^\top$ has no zero eigenvalues and therefore $\Q$ is full rank. This then proves that $\W_1\C_{xx}\W_1^\top=\I_k$ is satisfied at the optimum as desired. To proceed, we multiply  Eq.~\eqref{eq:Vx_update_apdx} by $\W_1^\top$ on the right  to get:
\begin{align}\label{eq:interim_relation}
    0&=\W_1 \C_{xy} \C_{yx}\W_1^\top-\Q\Q^\top\W_1\C_{xx}\W_1^\top.
\end{align}
If we plug this back into the objective \eqref{eq:VxQ_obj}, we see after cancellations that the only remaining term in the objective is $-\Q\Q^\top$. We will get back to this point below. We now use Eq.~\eqref{eq:Vx_update_apdx} and the relationship~\eqref{eq:interim_relation} to solve for $\Q\Q^\top$:
\begin{equation}\label{eq:commute1}
    \Q\Q^\top = \tilde\W_1 \C_{xx}^{-\frac12}\C_{xy} \C_{yx}\C_{xx}^{-\frac12}\tilde\W_1^\top(\tilde\W_1\tilde\W_1^\top)^{-1},
\end{equation}
where we have defined $\tilde \W_1:=\W_1\C_{xx}^\frac12$.
Since $\Q\Q^\top$ is symmetric, we can take the transpose of both sides of this equation to write:
\begin{equation}\label{eq:commute2}
    \Q\Q^\top = (\tilde\W_1\tilde\W_1^\top)^{-1}\tilde\W_1 \C_{xx}^{-\frac12}\C_{xy} \C_{yx}\C_{xx}^{-\frac12}\tilde\W_1^\top.
\end{equation}
Comparing Eq.~\eqref{eq:commute1} and Eq.~\eqref{eq:commute2}, we see that $(\tilde\W_1\tilde\W_1^\top)^{-1}$ and $\tilde\W_1 \C_{xx}^{-\frac12}\C_{xy} \C_{yx}\C_{xx}^{-\frac12}\tilde\W_1^\top$ commute. Therefore, they also commute with $(\tilde\W_1\tilde\W_1^\top)^{-1/2}$. We can use this to write $\Q\Q^\top$ as:
\begin{equation}
    \Q\Q^\top = \U \C_{xx}^{-\frac12}\C_{xy} \C_{yx}\C_{xx}^{-\frac12} \U^\top,
\end{equation}
where we have defined the semi-orthogonal matrix $\U=(\tilde\W_1\tilde\W_1^\top)^{-\frac12}\tilde\W_1$. 
Plugging everything back into the objective, and recalling that the only remaining term in the objective is $-\Q\Q^\top$ we get:
\begin{equation}
    \min_{\U\in\R^{k\times m}}\tr  -\U\C_{xx}^{-\frac12}\C_{xy}\C_{yx}\C_{xx}^{-\frac12}\U^\top
    \;\text{ such that }\; \U\U^\top = \I_k\label{eq:cca_final_apdx}.
\end{equation}
The minimum of this objective is when $\U$ aligns with the top $k$ eigenvectors of the matrix $\M:=\C_{xx}^{-\frac12}\C_{xy} \C_{yx}\C_{xx}^{-\frac12}$. As $\M = \F\F^\top$ with $\F:=\C_{xx}^{-\frac12}\C_{xy}$, the rank of $\M$ is equal to the rank of $\F$ which is equal to the rank of $\C_{xy}$. Therefore, if $\C_{xy}$ has at least $k$ non-zero eigenvalues, then at the optimum, $\Q\Q^\top$ has no zero eigenvalues and $\W_1\C_{xx}\W_1^\top=\I_k$, that is the inequality constraint is saturated, which we set out to show.

\section{Interpretation of the BMVR teaching signal}\label{app:bioRRR_interpretation}
We noted that the BMVR algorithm does not explicitly compute the loss or the loss gradient. However, the teaching signal for the first layer weight updates given in Eq.~\eqref{eq:W1_rule}:
\begin{align*}
    \W_1&\gets\W_1+\eta(\a_t-\Q\n_t)\x_t^\top
\end{align*}
can be interpreted as a backpropagated error signal near the equilibrium of the loss function. Comparing this update rule to that of backprop given in Eq.~\eqref{eq:W1_NN_rule}:
\begin{equation*}
    \delta \W_1 \propto (\W_2^\top \epsilon_t)\x_t^\top \quad,\quad \epsilon_t = (\y_t-\hat \y_t),
\end{equation*}
we see that the backpropagated error $(\W_2^\top \epsilon_t)$ in the backprop algorithm is replaced by the term $(\a_t-\Q\n_t)$ in BMVR. In Sec.~\ref{sec:circuit} we argued that this signal resembles the calcium plateau potential in cortical microcircuits. Here, we show that near the equilibrium, this term also carries a backpropagated error signal similar but not equal to $(\W_2^\top \epsilon_t)$ in the backprop algorithm.

To proceed, we look at the optimum of the objective where, from Eq.~\eqref{eq:Vx_update_apdx}, we have
\begin{equation*}
    \Q\Q^\top\W_1 = \W_2\C_{yx}\C_{xx}^{-1}\;
    \;\Rightarrow\;
    \Q\n_t = \Q\Q^\top\W_1 \x_t =  \W_2^\top\C_{yx}\C_{xx}^{-1} \x_t = \W_2^\top \tilde\y_t,
\end{equation*}
where we have used $\n_t=\Q^\top\z_t$ and $\z_t=\W_1\x_t$. Here, we have defined $\tilde\y_t := \C_{yx}\C_{xx}^{-1} \x_t$.
As \mbox{$\C_{yx} \C_{xx}^{-1}=\text{arg$\,$min}_\W \lVert\Y-\W\X\rVert^2_{\Sig}$} is the optimum of a rank-unconstrained regression objective, $\tilde\y_t$ is the best estimate of $\y_t$ given the samples received thus far. This quantity is different from $\hat\y_t = \W_2\W_1\x_t$ which is the best estimate of a rank-constrained objective of $\y_t$ at equilibrium. These two would be equivalent only if $k>\min(m,n)$.

Using these quantities and the definition of $\a_t = \W_2^\top \y_t$, we can rewrite the quantity $\a_t-\Q\n_t$ and the $\W_1$ update in Eq.~\eqref{eq:W1_rule} as
\begin{align}
    \a_t - \Q\n_t = \W_2^\top (\y_t-\tilde\y_t)\quad\Rightarrow\quad
    \W_1&\gets\W_1+\eta\;\Big[\W_2^\top(\!\underbrace{\,\y_t-\tilde\y_t}_{\text{prediction error}}\!)\,\Big]\x_t^\top \label{eq:Vx_at_equil}.
\end{align}
Therefore, while the error term $\y_t-\tilde\y_t$ and backpropagation are not present explicitly in BMVR, at the optimum, the teaching signal $\a_t-\Q\n_t$ is equal to a backpropagated error signal, and the update of $\W_1$ is proportional to the covariance of this backpropagated error signal and the input $\x_t^\top$.

\section{Comparison with prior work}\label{app:prior_work}

We previously discussed the relationship between BMVR and backprop. Here we look at some of the similarities and differences with other solutions proposed in response to the biological plausibility problems of backprop. The algorithms we compare to have the benefit that they are applicable to deep neural networks while BMVR as presented in this paper is only available for a 2-layer neural network. Here we compare all algorithms as they would be applied to this 2-layer network and leave the comparison in the case of deeper networks to future work.

The BMVR algorithm bears strong resemblance to the Target Propagation (TP) algorithm~\cite{Lee2014} in the sense that in both algorithms the backpropagated quantity is the target (or the label) and not a computed error. However, there are major differences between these algorithms. In TP, there is a reconstruction loss added to the objective to make sure that the forward and backward weights are in effect the inverses of each other. This is not necessary in BMVR.\footnote{In fact the forward weights of the second layer are not present in the BMVR algorithm and play no role in the training of the first layer weights. These can be included without any restriction.} Furthermore the TP algorithm does not account for the presence or the utility of the interneurons of this microcircuit.\footnote{The differences are even greater in Difference Target Propagation where both the target \emph{and} the putative forward quantity are both backpropagated independently.} Overall, the BMVR algorithm follows the spirit of target propagation in that the backpropagated quantity is the target and not the error but differs greatly from TP in implementation and biological realism.

The role of the calcium plateau potential as a teaching signal was previously noted in experimental work~\cite{golding2002,bittner2015,bittner2017behavioral,GreinbergerMagee2020}. Based on these observations, empirical models of cortical circuits have been proposed~\cite{Urbanczik2014,Sacramento2018,Milstein2020}. In~\cite{Sacramento2018}, it was shown that a deep network based on calcium plateau-like teaching signals can approximate the backpropagation algorithm in certain limits. However, the model of~\cite{Sacramento2018} is limited in part because of its requirement of direct one-to-one feedback connections between the pyramidal neurons of each layer and the interneurons of the previous layer. In the BMVR algorithm, these connections are not required.

\section{Decoupling the interneuron synapses}\label{app:decoupled_weights}

The BMVR neural circuit derived in Sec.~\ref{sec:circuit}, with  learning rules given in Eqs.~\eqref{eq:W1_rule}$-$\eqref{eq:Q_rule}, requires the pyramidal-to-interneuron weight matrix~($\Q^\top$) to be the the transpose of the interneuron-to-pyramidal weight matrix~($\Q$). Naively, this is not biologically plausible and is another example of the weight transport problem of backprop, albeit a less severe one as both sets of neurons (pyramidal and interneurons) are roughly in the same region of the brain. Here, we show that the symmetry between these two sets of weights ($\Q$ and $\Q^\top$) follows from the operation of local learning rules. The argument is similar to that of the supervised predictive coding network discussed in~\cite{Whittington2017}.

\begin{figure}
\centering
\includegraphics[width=\textwidth]{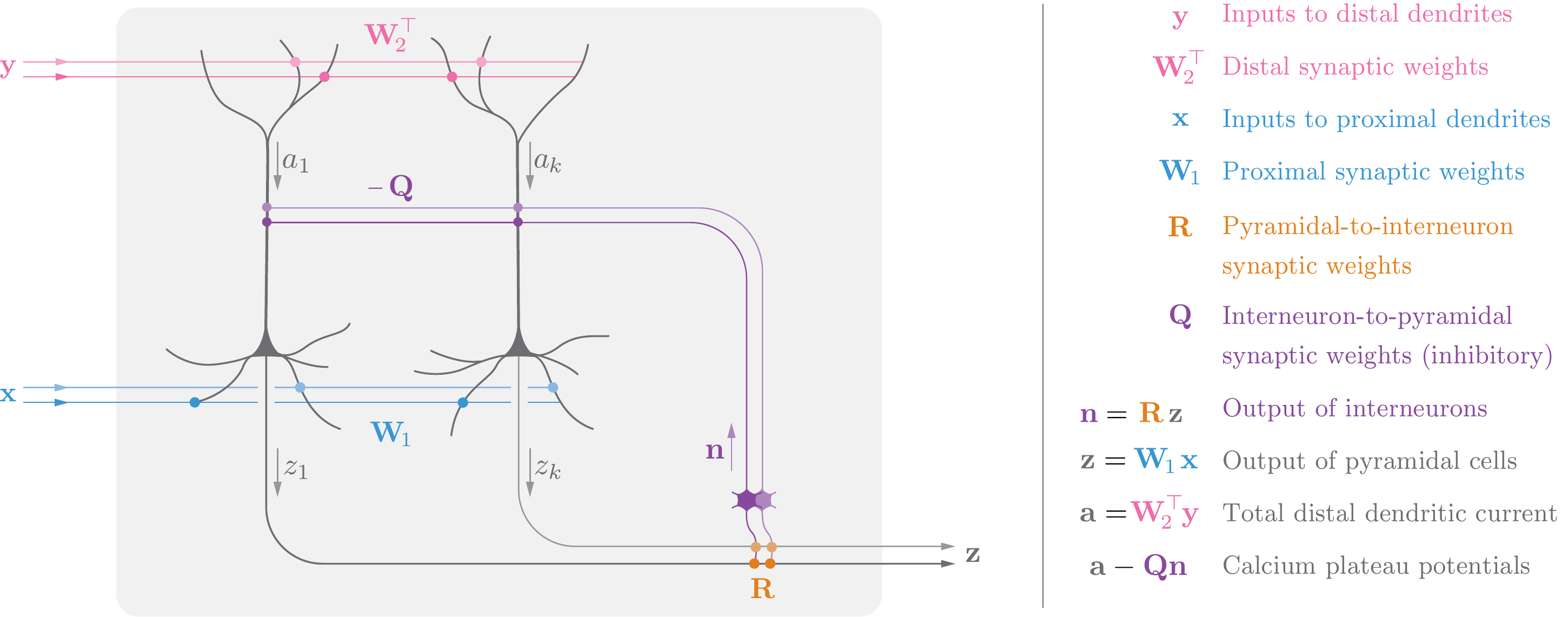}
\caption{\small The BMVR circuit with decoupled interneuron-to-pyramidal weights ($\Q$) and pyramidal-to-interneuron weights ($\bf R$). Following Hebbian learning rules, the weights $\bf R$ approach $\Q^\top$ exponentially.}
\label{fig:deoupled_microcircuit}
\end{figure}

To derive fully biologically plausible learning rules, we replace the pyramidal-to-interneuron weight matrix~($\Q^\top$) by a new weight matrix $\bf R$ which a~priori is unrelated to $\Q$ (Fig.~\ref{fig:deoupled_microcircuit}). We then impose the Hebbian learning rules for both sets of weights
\begin{align}
    \Q \gets&\Q+\frac{\eta}{\tau}(\z_t\n_t^\top-\Q) \\
    \bf R \gets&\bf R+\frac{\eta}{\tau}(\n_t\z_t^\top-\bf R).
\end{align}
If we assume that $\Q$ and $\bf R$ assume values $\Q_0$ and $\bf R_0$ at time $t=0$, after viewing $T$ samples, the difference $\Q^\top - \bf R$ can be written in terms of the initial values as
\begin{equation}
\Q^\top - {\bf R} = (1-\eta/\tau)^T (\Q_0^\top-\bf R_0).
\end{equation}
We see that the difference decays exponentially. Therefore, after viewing a finite number of samples, $\bf R$ would be approximately equal to $\Q^\top$ and we get back the BMVR update rules.

\section{Numerical experiment details} \label{app:exp_details}
In this section we provide the hyperparameters for the numerical experiments of Sec.~\ref{sec:experiments}. We note that because of the biophysical differences between the apical and basal synapses as well as the interneuron synapses, the learning rate of these synaptic efficacies is not necessarily the same. For this purpose, in our experiments we allow for different learning rates for $\W_1$, $\W_2$ and $\Q$. In all cases we perform a coarse hyperparameter search between 1 and 0.0001 with a logarithmic grid containing 12 data points (3 per decade). To keep the comparison between BMVR and backprop fair, we perform a similar decoupled hyperparameter search for $\W_1$ and $\W_2$. For each quantity, we also allow for a learning rate decay of the form $\eta=\frac{\eta_0}{1+t/t_0}$, where $t_0$ is another hyperparameter determined via a grid search.

For the linear experiments with results in Fig.~\ref{fig:experiment}, the hyperparameters used in the experiment are given in Tab.~\ref{tab:linear_hyperparameters}. For the non-linear experiments results reported in Tab.~\ref{tab:nn}, we pick a fixed learning rate without decay. The non-linearity chosen is a mean-subtracted ReLU, where the mean of the neuron is estimated in an online manner:
\begin{equation}
    \bar \z = \bar \z + \epsilon (\z_t-\bar\z)
\end{equation}
In our experiments we take $\epsilon = 10^{-4}$. For increased difficulty on the task we train the network on the 10,000 ``test'' samples of the MNIST dataset and then test it on the 50,000 ``train'' samples. The hyperparameters of this experiment are given in Tab.~\ref{tab:nn_hyperparameters}.

\renewcommand{\arraystretch}{1.5}
\begin{table}
\centering
  \begin{tabular}{cccccc}
    \toprule
        &   \multicolumn{3}{c}{BMVR}  &    \multicolumn{2}{c}{Backprop}     \\
        &   $\eta_{w_1}$    &   $\eta_{w_2}$    &   $\eta_{q}$ &   $\eta_{w_1}$    &   $\eta_{w_2}$\\
    \midrule
    MNIST   &   $\frac{0.01}{1+t/10^3}$   &   $\frac{0.01}{1+t/10^3}$   &   $\frac{0.003}{1+t/10^3}$
            &   $\frac{0.02}{1+t/10^3}$    &    $\frac{0.02}{1+t/10^3}$   \\
    FMNIST   &   $\frac{0.013}{1+t/10^3}$   &   $\frac{0.013}{1+t/10^3}$   &   $\frac{0.005}{1+t/10^3}$ 
            &   $\frac{0.018}{1+t/10^3}$    &    $\frac{0.018}{1+t/10^3}$   \\
    CIFAR-10 &   $\frac{0.01}{1+t/1.5\times10^4}$   &   $\frac{0.002}{1+t/1.5\times10^4}$   &   $\frac{0.002}{1+t/1.5\times10^4}$ 
            &   $\frac{0.0065}{1+t/10^4}$    &    $\frac{0.0065}{1+t/10^4}$   \\
    CIFAR-100 &   $\frac{0.025}{1+t/4\times10^4}$   &   $\frac{0.001}{1+t/4\times10^4}$   &   $\frac{0.002}{1+t/4\times10^4}$ 
            &   $\frac{0.0065}{1+t/1.1\times10^4}$    &    $\frac{0.0065}{1+t/1.1\times10^4}$   \\
    \bottomrule
  \end{tabular}\vspace{5pt}
\caption{Hyperparameter choices for the linear experiment with results reported in Fig.~\ref{fig:experiment}.}
\label{tab:linear_hyperparameters}
\end{table}

\begin{table}
\centering
  \begin{tabular}{cccccc}
    \toprule
        &   \multicolumn{3}{c}{BMVR}  &    \multicolumn{2}{c}{Backprop}     \\
        &   $\eta_{w_1}$    &   $\eta_{w_2}$    &   $\eta_{q}$ &   $\eta_{w_1}$    &   $\eta_{w_2}$\\
    \midrule
    $k=64$   &   0.001   &   0.0002   &   0.001
            &   $0.4$    &    $0.4$   \\
    $k=256$   &   0.2   &   0.04   &   0.04
            &   $0.2$    &    $0.2$   \\
    \bottomrule
  \end{tabular}\vspace{5pt}
\caption{Hyperparameter choices for the non-linear experiment with results reported in Tab.~\ref{tab:nn}.}
\label{tab:nn_hyperparameters}
\end{table}

\end{document}